%% file: main.tex
\documentclass[letterpaper,journal]{IEEEtran}
\usepackage[margin=1in,bottom=1.05in,includehead,headsep=12pt]{geometry}

\usepackage{graphicx}
\usepackage{booktabs}
\usepackage{multirow}
\usepackage{multicol}
\usepackage{array}
\usepackage{tabularx}
\usepackage[table]{xcolor}
\usepackage{amsmath}
\usepackage{amsfonts}
\usepackage{amssymb}
\usepackage{algorithm}
\usepackage{algpseudocode}
\usepackage{placeins}
\usepackage{listings}
\usepackage{stfloats}
\usepackage{float}
\usepackage{needspace}
\usepackage{textcomp}
\usepackage{url}
\usepackage{cite}
\usepackage[hidelinks]{hyperref}

\newcolumntype{Y}{>{\centering\arraybackslash}X}

\newcommand{\method}{PIVOT}
\newcommand{\bench}{PhysForensics-Bench}

\lstdefinestyle{promptbox}{
  basicstyle=\ttfamily\fontsize{5.8pt}{6.6pt}\selectfont,
  breaklines=true,
  breakatwhitespace=false,
  columns=fullflexible,
  keepspaces=true,
  frame=single,
  framesep=3pt,
  aboveskip=7pt,
  belowskip=9pt,
  framerule=0.35pt,
  rulecolor=\color{black!35},
  backgroundcolor=\color{black!2},
  xleftmargin=4pt,
  xrightmargin=4pt
}
\lstdefinestyle{jsonbox}{
  style=promptbox,
  language={},
  showstringspaces=false
}
\newcommand{\papertitle}{PIVOT: Physics-Grounded Verification for AI-Generated Audio-Video Detection}
\newcommand{\exampleheading}[1]{\par\medskip\noindent{\small\bfseries #1}\par\nobreak\smallskip}

\title{\papertitle}
\author{Bo~Zheng, Kangran~Zhao, Xiaoyu~Zhang, Weinan~Guan, Zhiheng~Li,
Yize~Chen, Haizhou~Li, Qingshan~Liu, Siwei Lyu, and~Baoyuan~Wu%
\thanks{Bo Zheng, Kangran Zhao, Xiaoyu Zhang, Weinan Guan,
Zhiheng Li, Yize Chen, Haizhou Li, and Baoyuan Wu are with the School of Artificial Intelligence, Chinese University of Hong Kong, Shenzhen, Shenzhen 518172, China. 
Qingshan Liu is with Nanjing University of Posts and
Telecommunications, Nanjing 210023, China.
Siwei Lyu is with University at Buffalo, State University of New York, New York 14260, USA.}
\thanks{Corresponding author: Baoyuan Wu (wubaoyuan@cuhk.edu.cn).}}
\begin{document}
\maketitle
\suppressfloats[t]

\begin{figure}[t]
  \centering
  \includegraphics[width=\columnwidth]{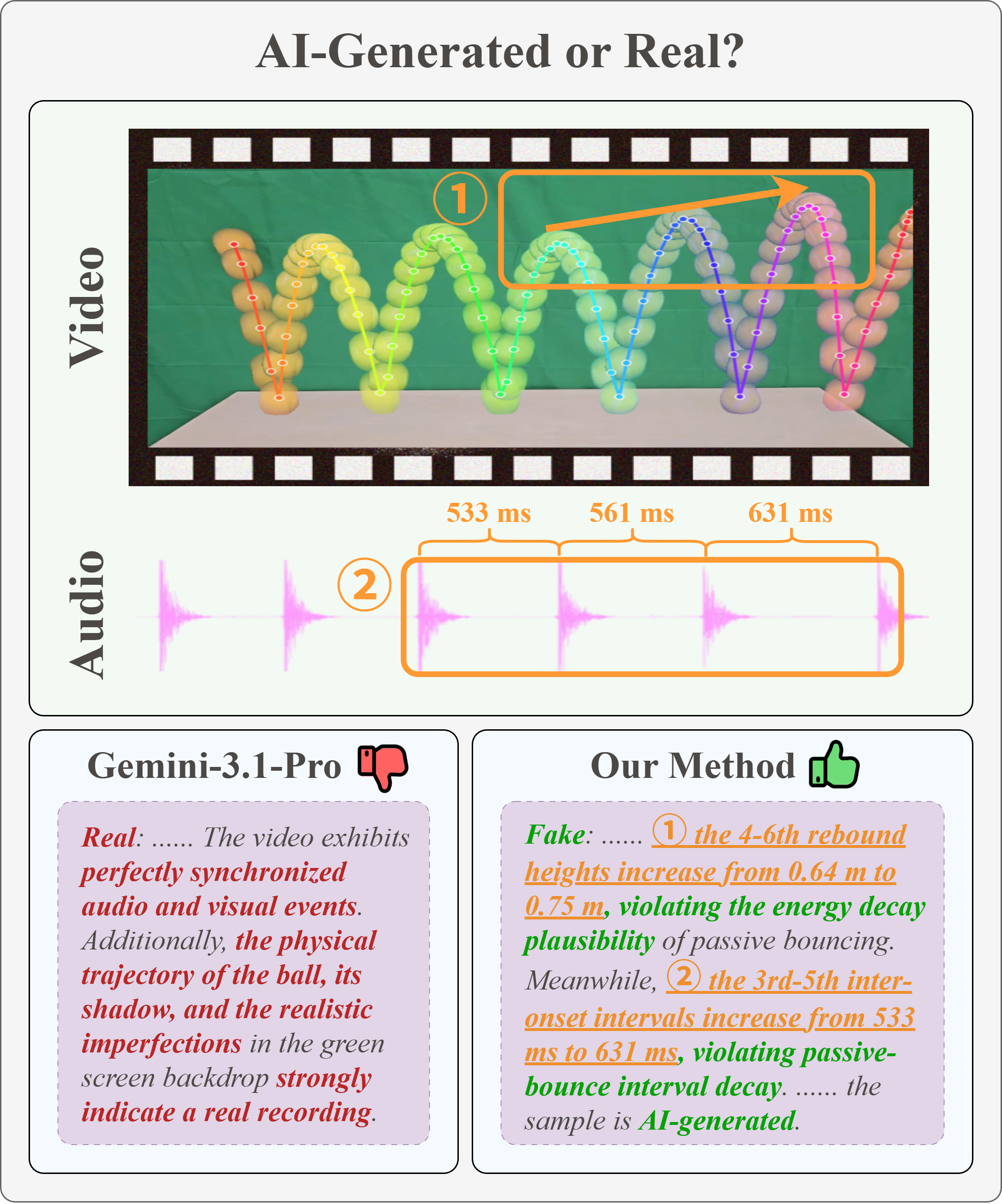}
  \caption{\textbf{From artifact cues to physics-grounded verification.}
  Direct foundation-model inspection is persuaded by surface realism, whereas
  \method{} identifies physically implausible increases in rebound height and
  audio inter-onset interval and supports its AI-generated decision with
  evidence of physical inconsistency.}
  \label{fig:motivation}
\end{figure}

\input{sections/abstract}

\input{sections/introduction}

\input{sections/related_work}

\input{sections/method}
\input{sections/physav_bench}

\input{sections/experiments}

\input{sections/limitations}

\input{sections/conclusion}

\clearpage
\appendix
\input{sections/appendix}

\clearpage
\onecolumn
\begin{multicols}{2}
\bibliographystyle{IEEEtran}
\bibliography{references_new}
\end{multicols}

\end{document}

%% file: sections/abstract.tex
\begin{abstract}
As generative models continue to advance, AI-generated content (AIGC) is becoming increasingly realistic, weakening the artifact cues commonly exploited by existing detectors. Nevertheless, faithfully reproducing the physical behavior of real-world events remains challenging for current generators. We therefore explore detecting AIGC by assessing whether the depicted event satisfies measurable constraints derived from physical laws. 
We introduce \method{}, a physics-grounded AIGC detector, instantiated here for audio-video clips, that estimates physical quantities from video and
audio, selects physical laws relevant to each clip, and verifies their
measurable constraints.
Beyond a real/fake decision, \method{} returns supporting evidence that records the verification outcome, relevant time window, and supporting quantities for each applicable law. 
Although instantiated and evaluated here on audio-video data, the framework can, in principle, extend to other AIGC modalities whenever the physical quantities required for verification can be estimated reliably.
We also introduce \bench{}, comprising paired real and generated audio-video clips from nine event-centric scene families and two recent audio-video generators. On \bench{}, \method{} achieves 70.30\% accuracy and 64.29\% F1 score on Real+Seedance, and 72.16\% accuracy and 65.82\% F1 on Real+VEO. In comparison, direct inspection with Gemini 3.1 Pro obtains 53.96\% accuracy and 60.09\% F1 on Real+Seedance, and 57.22\% accuracy and 63.44\% F1 on Real+Veo. These results demonstrate the practical promise of physical-consistency verification as a structured and inspectable source of evidence that complements artifact-based AIGC detection.
\end{abstract}


%% file: sections/introduction.tex
\section{Introduction}
\label{sec:introduction}

\IEEEPARstart{A}{udio-video} generation is rapidly advancing toward clips with
realistic appearance and coherent motion~\cite{singer2022make,bar2024lumiere,sora2024},
while recent systems also generate synchronized sound~\cite{polyak2024movie,seedance2026seedance,veo2026}.
As such media become easier to deploy in communication settings, fabricated
events can more readily circulate as apparent recordings, making reliable
authenticity assessment increasingly important~\cite{verdoliva2020media,mirsky2021creation}.
Existing detection research already provides valuable signals from visual
appearance, frame consistency, and temporal dynamics~\cite{ma2025detecting,zheng2025d3,zhang2026physics},
as well as audio-visual correspondence and local cross-modal
inconsistencies~\cite{oorloff2024avff,astrid2025audio,shahzad2026save,wei2026generalizing}.
However, some model-exposed artifact cues may become weaker or less stable as
the generators evolve, as cross-generator evidence from image forensics
suggests~\cite{wang2020cnn,ojha2023towards}. This motivates the search for more
durable evidence grounded in the properties of the depicted event itself.

Physical laws offer a promising basis for such evidence because perceptual
realism does not guaranty physical fidelity. Recent studies show
that visually convincing generated videos can still violate physical
commonsense in object interactions, material behavior, and action-centric
events~\cite{bansal2025videophy,meng2024towards,bansal2025videophy2}. The similar
challenge also exists in joint audio-video generation:
a clip may look and sound plausible at first glance while remaining
inconsistent in visual dynamics, acoustic behavior, or the physical
relationship between an event and its sound~\cite{cui2026joint,xie2025phyavbench}.
These findings suggest that physical laws can guide the interpretation of
observations as evidence about the depicted event, complementing artifact cues
whose reliability may change as generation techniques evolve.

Motivated by this gap, we formulate audio-video AIGC detection as
physics-grounded verification: testing whether the depicted event is consistent with measurable constraints derived from physical laws. 
Audio-visual studies show that visible events can be associated with acoustic
events and sound-producing objects from paired observations~\cite{arandjelovic2017look,arandjelovic2018objects,owens2018audio}.
Besides, physical reasoning research has likewise studied object dynamics and latent physical properties from visual observations, as well as physical reasoning in simulated environments~\cite{battaglia2013simulation,mottaghi2016newtonian,wu2016physics,bakhtin2019phyre}.
Together, these lines of research suggest that audio-video observations can 
expose physical quantities and cross-modal relations useful for law-based consistency 
assessment. Given the depicted event and the available quantity estimates, a physical 
law that is relevant to the event and verifiable from the available quantities can be 
operationalized as one or more measurable constraints. The estimated quantities
provide evidence supporting or contradicting these constraints. In principle, this
formulation can extend to additional families of physical laws as reliable estimators for 
their required quantities become available. In the current instantiation, we focus on 
event-centric clips whose observable physical evidence is primarily derived from object 
motion, contact dynamics, scene geometry, and associated acoustic events.

Figure~\ref{fig:motivation} illustrates this verification perspective. In the example, direct
inspection with Gemini 3.1 Pro~\cite{google2026gemini31pro} classifies the clip as real based on the
apparently synchronized audio and visual events, plausible ball trajectory and
shadow, and realistic imperfections in the backdrop. Yet the measured event
contains two physically implausible trends. First, the fourth through sixth
rebound heights increase from $0.64\,\mathrm{m}$ to $0.75\,\mathrm{m}$ during
passive bouncing (\textcircled{1}), violating the expected decay of mechanical
energy. Second, the third through fifth audio inter-onset intervals increase
from $533\,\mathrm{ms}$ to $631\,\mathrm{ms}$ (\textcircled{2}), whereas
successive impact intervals should shorten as a passive bouncing system loses
energy. Rather than relying on an overall impression of realism, the proposed method \method{}
links these measurements to energy-decay and bounce-interval constraints and
returns the corresponding laws, time windows, verification statuses, and
supporting quantities as evidence for its decision.

We operationalize this formulation in \method{}. Given a clip, \method{}
estimates structured physical quantities from the video and audio streams,
selects laws that are relevant to the depicted event and measurable from the
available quantity fields, and instantiates them as operational constraints. It
then evaluates each applicable constraint as supported, violated, or uncertain
and produces a real/fake decision together with supporting evidence organized
by law and time window; non-applicable law-window pairs are omitted. This
caption--estimate--select--verify design separates the observations, the
physically expected behavior, and the verification outcome while allowing
stronger physical estimators to be incorporated through the same quantity
interface.

To evaluate this direction, we introduce \bench{}, a physics-centric benchmark
of paired real and generated audio-video clips spanning nine event-centric
scene families and two recent generators, Veo~3.1 Fast~\cite{veo2026} and Seedance 2.0~\cite{seedance2026seedance}. On the test split of \bench{},
\method{} achieves 70.30\% accuracy and 64.29\% F1 score on
Real+Seedance, and 72.16\% accuracy and 65.82\% F1 score on Real+Veo.
In comparison, direct inspection with Gemini 3.1 Pro obtains 53.96\% accuracy
and 60.09\% F1 score on Real+Seedance, and 57.22\% accuracy and
63.44\% F1 score on Real+Veo.

Our contributions are threefold:
\begin{itemize}
    \item We formulate audio-video AIGC detection as physics-grounded
    verification and instantiate this view in \method{}, which selects
    event-relevant and measurable physical laws, verifies their constraints over estimated
    quantities, and returns supporting evidence organized by law and time
    window.

    \item We build \bench{}, a physics-centric benchmark of paired real and
    generated audio-video clips spanning nine event-centric scene families and
    two recent generators.

    \item We demonstrate the practical potential of physical-consistency verification
    on \bench{} through comparisons with existing detectors and direct
    foundation-model inspection, together with ablations of the video and audio
    physical estimators.
\end{itemize}

%% file: sections/related_work.tex
\section{Related Work}

\begin{figure*}[t]
  \centering
  \includegraphics[width=\textwidth]{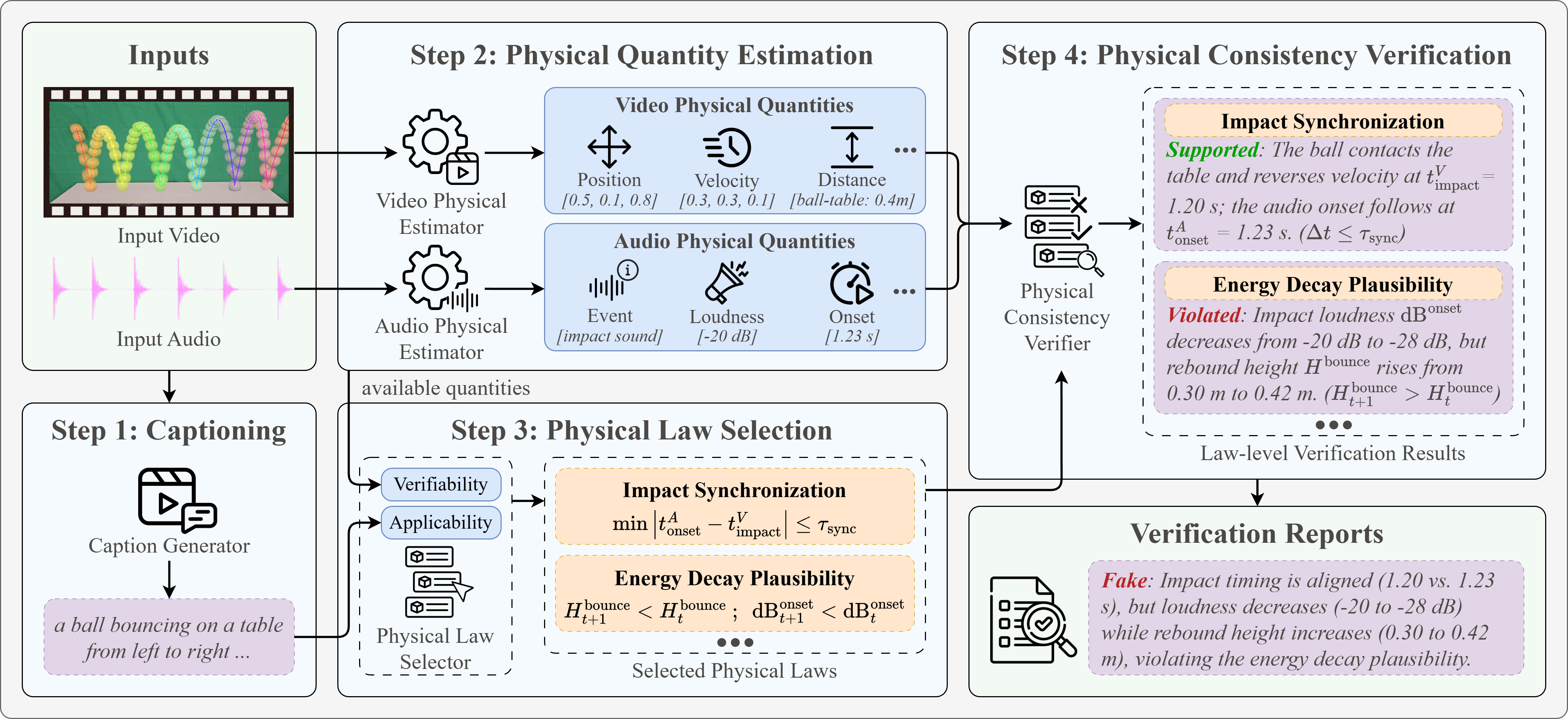}
  \caption{\method{} inference pipeline. \textbf{Step 1: Captioning} generates
  clip captions without making a detection decision. \textbf{Step 2: Physical
  Quantity Estimation} extracts time-indexed video and audio quantities.
  \textbf{Step 3: Physical Law Selection} selects laws that are relevant to the
  depicted event and verifiable from the available quantities, then instantiates
  their physical constraints.
  \textbf{Step 4: Physical Consistency Verification} evaluates those constraints against the
  quantities and returns a real/fake decision with supporting evidence.}
  \label{fig:pipeline}
\end{figure*}

\paragraph{AI-generated video detection}
Video forensics has been widely studied in face manipulation and deepfake
settings. MesoNet uses compact convolutional networks to detect facial video
forgeries~\cite{afchar2018mesonet}. FaceForensics++ establishes a large-scale
benchmark and dataset for manipulated-face detection
~\cite{rossler2019faceforensics++}; Celeb-DF contributes high-quality celebrity
DeepFake videos~\cite{li2020celeb}; and DFDC provides a large face-swap video
dataset~\cite{dolhansky2020deepfake}. Beyond these evaluation resources, F3-Net
mines frequency-aware forgery clues~\cite{qian2020thinking}, while LipForensics
detects semantic irregularities in mouth motion using representations learned
through visual-speech recognition~\cite{haliassos2021lips}.
X2-DFD combines an MLLM with specialized
forgery-feature detectors to produce both deepfake decisions and human-readable
explanations~\cite{chen2025x2dfd}.

Recent work also studies general AI-generated content detection beyond face
manipulation. SIDA distinguishes authentic, fully synthetic, and tampered
social-media content, while localizing tampered regions and providing textual
explanations~\cite{huang2025sida}. So-Fake provides a large-scale,
generator-diverse benchmark with dedicated out-of-domain evaluation using
generators excluded from training~\cite{huang2025sofake}.
DeCoF learns temporal artifacts through frame consistency
~\cite{ma2025detecting}. DeMamba introduces the million-scale GenVideo benchmark
and a plug-and-play module for spatial-temporal inconsistency analysis
~\cite{chen2026demamba}. GenVidBench further provides 6.78 million videos with
cross-source and cross-generator evaluation settings~\cite{ni2026genvidbench}.
Among existing detectors, D3~\cite{zheng2025d3} and NSG-VD~\cite{zhang2026physics} are particularly relevant to our physics-grounded perspective. D3 derives second-order temporal features motivated by Newtonian mechanics and uses their standard deviation as its detection score~\cite{zheng2025d3}. NSG-VD derives a Normalized Spatiotemporal
Gradient from probability-flow conservation and uses deep-kernel MMD to compare
the NSG features of a test video with those of reference real videos as its
detection metric~\cite{zhang2026physics}. These methods demonstrate the value of
physics-inspired temporal priors for AI-generated video detection. Their
published inference pipelines apply a predefined physics-inspired formulation
across inputs, rather than explicitly determining which physical laws are
relevant to each depicted event and verifiable from its available quantities.
This distinction motivates the event-aware, measurability-constrained
physical-consistency verification setting studied in our method.

\paragraph{AI-generated audio-video detection}
Audio-visual fake media detection has also been studied in multimodal deepfake
settings. FakeAVCeleb provides a multimodal dataset containing synthesized faces
and voices~\cite{khalid2021fakeavceleb}, while LAV-DF introduces localized audio,
visual, and audio-visual manipulations for detection and temporal localization
~\cite{cai2023glitch}. AVFF learns audio-visual correspondence through
self-supervised representation learning followed by supervised deepfake
classification~\cite{oorloff2024avff}. FGI targets fine-grained spatial and
temporal audio-visual inconsistencies~\cite{astrid2024detecting}. Astrid et al.
further use temporal distance maps and locally inconsistent pseudo-fakes to
capture local temporal mismatches~\cite{astrid2025audio}. SAVe combines facial
manipulation cues with lip-speech misalignment while learning from authentic
videos~\cite{shahzad2026save}, whereas AVPF generates audio-visual pseudo-fakes
from authentic samples to improve generalization~\cite{wei2026generalizing}.
DeepfakeBench-MM provides a unified evaluation framework spanning five
multimodal datasets and eleven detectors, together with the Mega-MMDF dataset of
human-centric audio-visual forgeries~\cite{zhao2025deepfakebenchmm}.

General audio-video AIGC detection remains less explored than face-centric
audio-visual deepfake detection. MVAD broadens multimodal AIGC detection data
beyond facial deepfakes~\cite{hu2025mvad}. AV-Phys Bench evaluates physical
commonsense in joint audio-video generation and introduces AV-Phys Agent, which
combines multimodal reasoning with deterministic acoustic measurement tools for
rubric-based generation evaluation~\cite{cui2026joint}. PhyAVBench evaluates the
audio-physics sensitivity of text-to-audio-video generators through controlled
paired prompts~\cite{xie2025phyavbench}. These studies expand general
audio-video data and physics-aware generation evaluation, while most of the
audio-visual detectors considered above are designed for facial manipulation,
lip-speech alignment, or local cross-modal inconsistency. Authenticity detection
of general physical events using per-law measurable evidence remains less
studied.

\paragraph{Physical reasoning for perception}
Physical reasoning has long supported scene understanding. Approximate physical
simulation can explain human judgments about scene dynamics
~\cite{battaglia2013simulation}, and Newtonian scene understanding predicts
forces and long-term object motion from static images
~\cite{mottaghi2016newtonian}. Physics 101 learns object properties from
unlabeled videos by encoding physical laws~\cite{wu2016physics}, while PHYRE
provides a benchmark of classical-mechanics puzzles for physical reasoning
~\cite{bakhtin2019phyre}.

Audio-visual learning provides complementary foundations. Visually Indicated
Sounds predicts impact sounds from silent videos to capture material properties
and physical interactions~\cite{owens2016visually}. Look, Listen and Learn
introduces audio-visual correspondence as a self-supervised learning task
~\cite{arandjelovic2017look}; Objects that Sound uses this correspondence for
cross-modal retrieval and sound-source localization
~\cite{arandjelovic2018objects}; and multisensory scene analysis learns from the
temporal alignment of video and audio~\cite{owens2018audio}.

Recent benchmarks directly evaluate physical commonsense in generated videos.
VideoPhy evaluates physical plausibility in text-to-video outputs
~\cite{bansal2025videophy}; PhyGenBench organizes evaluation around explicit
physical laws~\cite{meng2024towards}; and VideoPhy-2 extends this direction to
action-centric physical commonsense~\cite{bansal2025videophy2}. Broader
video-generation evaluation is complementary: VBench decomposes generation
quality into fine-grained dimensions~\cite{huang2024vbench}, EvalCrafter covers
visual, content, motion, and text-video alignment quality
~\cite{liu2024evalcrafter}, and FETV adds multi-aspect and temporal-aware
evaluation of open-domain text-to-video generation~\cite{liu2023fetv}.

Collectively, these studies provide complementary foundations in generated-video
detection, audio-visual forensic analysis, physical perception, and physics-aware
generation evaluation. \method{} brings these directions together for
authenticity detection: it selects physical laws according to the depicted event
and the quantities available for measurement, instantiates their operational
constraints, and returns physical-consistency verification records organized by law,
time window, and estimated quantity.

%% file: sections/method.tex
\section{Methodology}
\label{sec:method}

\method{} detects generated audio-video by testing whether the observed event is
physically consistent. The inference pipeline in Figure~\ref{fig:pipeline}
comprises four stages: \emph{captioning} generates clip captions;
\emph{physical quantity estimation} recovers time-indexed video and audio
quantities; \emph{physical law selection} selects event-relevant and measurable laws and
instantiates their constraints; and \emph{physical consistency verification}
evaluates those constraints and produces a structured report. This decomposition
is the core framework contribution: physics-grounded AIGC detection should make
the observed event and its physical quantities explicit, state which physical
laws apply, and check the observed quantities against their constraints while retaining a structured decision
trace. The resulting inference workflow is caption--estimate--select--verify.

\begin{figure*}[t]
  \centering
  \includegraphics[width=\textwidth]{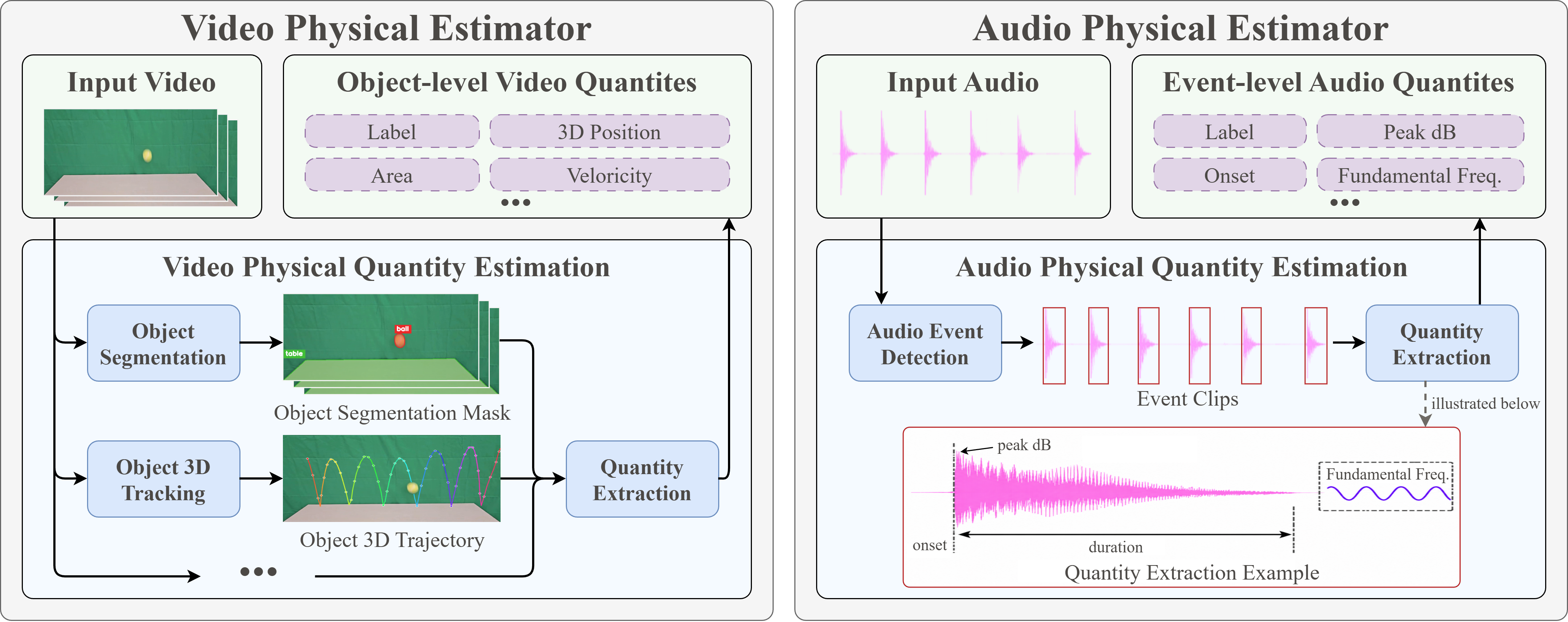}
  \caption{Illustrative implementations of the tool-agnostic physical quantity
  interface. \textbf{Top:} Basic visual attributes and kinematics are two
  example video branches; the latter illustrates one path for estimating
  position, velocity, and acceleration. External estimators may replace or
  extend either branch to expose other physical quantities.
  \textbf{Bottom:} The audio estimator localizes complementary sound events and
  spectral transients, fuses their temporal intervals, and produces event-level
  quantities such as onset, peak dB, F0, RMS, and HNR. The downstream framework
  consumes these structured quantities and does not depend on a particular
  estimator architecture.}
  \label{fig:physical-estimators}
\end{figure*}

\subsection{Captioning}

The captioning stage generates concise captions for the input clip, collectively
denoted by \(B_i\):
\begin{equation}
B_i = \mathcal{B}_{\theta}(\mathbf{x}_i).
\label{eq:clip-caption}
\end{equation}
It identifies the objects, surfaces, interactions, and temporal phases needed
to determine which physical laws are relevant, without predicting whether the
clip is real or generated. In the reported implementation, Gemini 3.1 Pro and
GPT-5.5 independently describe the clip, and both descriptions are retained in
\(B_i\) for downstream physical law selection.

\subsection{Physical Quantity Estimation}

\method{} treats video and audio physical estimators as external, replaceable
modules. Rather than prescribing a particular estimator architecture, the
framework requires only that each module expose named, time-aligned physical
quantities that can be inspected by later stages. For a clip \(\mathbf{x}_i\), we write
this shared interface as
\begin{equation}
M_i=(A_i,V_i).
\label{eq:quantities}
\end{equation}
Here \(A_i\) contains event-level audio records, and \(V_i\) contains
frame-level video records. Quantity fields retain their timestamps and, where
applicable, units and coordinate conventions. Downstream components consume
only these structured records, not estimator-specific hidden features. This
interface allows improved or domain-specific estimation tools to be substituted
without changing physical law selection or verification.

\subsubsection{Video Physical Estimator}

The video estimator interface can connect external models or procedures that
provide the physical quantities required by the framework. The two branches at
the top of Figure~\ref{fig:physical-estimators} are illustrative implementation
examples rather than a fixed or exhaustive decomposition. The \emph{basic
visual attributes} branch applies object segmentation to the video frames,
producing object labels and mask extents from which quantities such as visible
object area can be derived. The \emph{kinematics} branch applies 3D object
tracking to recover each object's trajectory in a common coordinate system,
from which position, velocity, and acceleration can be estimated. Other
external estimators can replace or extend these paths to provide object-pair
relations or additional physical quantities.

Together, these records form the frame-level set \(V_i\). The interface does
not require a particular segmentation, geometry, or tracking model; it requires
only consistent object identities, timestamps, field definitions, and coordinate
conventions. Our current instantiation focuses on basic visual quantities and kinematic quantities, while the
same interface can incorporate additional properties such as mass, volume,
density, elasticity, or material when suitable estimators become available.
Appendix~\ref{app:video-estimator} documents the concrete tools, calibration,
and reconstruction procedure used in our experiments.

\subsubsection{Audio Physical Estimator}

The audio estimator interface, shown at the bottom of
Figure~\ref{fig:physical-estimators}, represents the waveform as temporally
localized event records. An event--transient extractor combines complementary
cues: low-band event detection identifies semantically recognizable sound
regions, while high-band transient detection recovers brief spectral changes
that may not receive a reliable semantic label. Their temporal intervals are
fused before physical quantities are extracted, so both cues are described on a
shared time axis aligned with the video.

For each detected interval, the resulting audio record can include the onset,
envelope boundaries, duration, peak dB, RMS level, fundamental frequency (F0),
harmonic-to-noise ratio (HNR), and spectral descriptors. These records form the
event-level set \(A_i\). As with the video interface, \method{}
depends on the fields and their temporal meaning rather than on a particular
detector or acoustic backbone. Appendix~\ref{app:audio-evidence} specifies the
concrete audio tools, frequency bands, fusion rule, and thresholds used in our
experiments.

Because full frame-level quantity sequences are unnecessarily long for
prompt-based reasoning, \method{} applies a lightweight, caption-conditioned
compression step before law selection and verification:
\begin{equation}
\tilde M_i = \operatorname{Compress}(M_i;B_i).
\label{eq:compression}
\end{equation}
The compressed representation preserves temporally informative quantity records
and includes the metadata needed to interpret the retained rows. The concrete
sampling strategies and serialization procedure are implementation details 
described in Appendix~\ref{app:physical-aware-sampling}.

\subsection{Physical Law Selection}

The space of physical laws governing real-world events is broad, making
exhaustive verification neither tractable nor useful.
A meaningful test must satisfy two conditions: the law must be relevant to the
event depicted in the clip, and its requirements must be covered by the
physical quantities available to the framework. The Physical Law Selector
therefore identifies applicable physical laws and specifies constraints that
can be checked using the available quantities. This event-aware,
measurability-constrained law selection uses the clip
captions, available quantity fields, and a structured law schema.
The available quantity specification is obtained from the field names in the
compact quantity tables. The Physical Law Selector receives the clip captions
\(B_i\) and these available fields, but it does not directly classify the
clip. Instead, it selects a compact set of laws that are relevant to the
depicted event and verifiable from the available quantities, then instantiates
their constraints.
Each selected law follows a constrained output schema containing its name,
relative priority, physical basis, required audio and video fields,
\texttt{decision\_basis}, applicability conditions, and failure conditions.
The captions supply clip context, while the field specification restricts
selection to laws that can be checked using the exposed quantities. Physical
law selection is therefore
\begin{equation}
L_i =
\mathcal{P}_{\theta}\!\left(B_i,
\operatorname{Fields}(\tilde M_i)\right).
\label{eq:law-selection}
\end{equation}
The \texttt{decision\_basis} states the equality, inequality, ordering, trend,
or temporal relation to be checked. Either modality's required-field list may
be empty when that modality is not needed.

As one example, in a bouncing or collision scene, the selector may produce an
\emph{impact synchronization} check. Its constraint requires an acoustic onset to
be temporally consistent with a nearby visible contact, release, rebound,
relation-distance extremum, velocity reversal, or acceleration peak. It
specifies the audio onset fields and video quantities that can support or
refute the relation. Applicability and failure conditions require the verifier
to account for low frame rate, occlusion, missed audio events, and merged
acoustic envelopes. During verification, the model compares measured
\texttt{onset\_ms} values with the available event quantities, then returns
support, violation, or uncertainty together with the quantities used in that
judgment. This example is illustrative rather than a fixed template;
additional checks associated with the selected laws are summarized in Appendix
Section~\ref{app:law-examples}.

The Physical Law Selector turns the clip captions and quantity specification into
an explicit physics test. A generic audio-visual classifier can silently rely
on any correlation in its representation; \method{} instead requires the model
to state the applicable physical laws and their constraints before the verifier
checks the compact quantities against those constraints.

\subsection{Physical Consistency Verification}

The Physical Consistency Verifier receives the selected laws, their physical constraints
over the required quantities, the clip captions, and compressed audio and video
quantities. For each law, it evaluates the constraint against the relevant
quantities and records whether the constraint is supported, violated, or uncertain.
The quantity observations supporting each judgment form the evidence reported
with the decision. The compressed
representation retains the timestamps, units, coordinate conventions, and
sampling metadata needed to interpret each row.
Here, verification denotes formula-guided consistency assessment conditioned
on estimated physical quantities, rather than exact physical simulation or
full system identification.

Each of three independent verification runs returns a binary label and a
structured report:
\begin{equation}
\left(\hat y_i^{(r)},R_i^{(r)}\right)
=\mathcal V_\theta\!\left(L_i,B_i,\tilde M_i\right),
\qquad r\in\{1,2,3\}.
\label{eq:explanation-output}
\end{equation}
For each applicable law and time window, \(R_i^{(r)}\) records
\emph{support}, \emph{violate}, or \emph{uncertain}, together with confidence,
the relevant quantity observations and values when available, and a textual
explanation. Inapplicable laws are omitted. The final sample-level label is the
majority vote over the three binary labels, while the final structured output
retains the reports from all three runs:
\begin{equation}
\hat y_i
= \operatorname{Mode}\!\left(
\hat y_i^{(1)},\hat y_i^{(2)},\hat y_i^{(3)}
\right), \qquad
R_i=\{R_i^{(r)}\}_{r=1}^{3}.
\label{eq:verification-vote}
\end{equation}
Confidence values are retained in the structured reports but are neither averaged
nor used to weight the votes. 
Each run-level label is not obtained by
thresholding a hand-written law score; it is produced by the verifier's
structured report, which ties a generated decision to concrete violations such
as a discontinuous visual trajectory, an implausible acoustic transient
sequence, an acoustic onset without corresponding contact, or a rebound rhythm
whose video events and audio timing disagree. 
Thus, \method{} produces a detection result and its supporting evidence in the same structured output.

The selector-produced law weights communicate relative law priority to the
verifier; they are not coefficients of a deterministic classifier. Weighted
support and violation rates are retained only as report diagnostics and do not
determine the run-level or final label. Each structured LLM verification call
still produces the law-window states and one run-level binary label. The
deterministic post-processing is the local majority vote over three such labels.
Algorithm~\ref{alg:physav-inference} summarizes the complete inference path.

\begin{algorithm}[t]
\caption{\method{} inference for audio-video clip \(\mathbf{x}_i\)}
\label{alg:physav-inference}
\footnotesize
\begin{algorithmic}[1]
\Require Clip \(\mathbf{x}_i\)
\Ensure Label \(\hat y_i\) and verification reports \(\{R_i^{(r)}\}_{r=1}^{3}\)
\State \(B_i \gets \mathcal B_\theta(\mathbf{x}_i)\)
\State \(M_i \gets \Call{PhysicalQuantities}{\mathbf{x}_i}\)
\State \(\tilde M_i \gets \Call{Compress}{M_i,B_i}\)
\State \(L_i \gets \mathcal P_\theta(B_i,\Call{Fields}{\tilde M_i})\)
\For{\(r=1,2,3\)}
  \State \((\hat y_i^{(r)},R_i^{(r)})\gets
  \mathcal V_\theta(L_i,B_i,\tilde M_i)\)
\EndFor
\State \(\hat y_i\gets\Call{Mode}{\hat y_i^{(1)},\hat y_i^{(2)},\hat y_i^{(3)}}\)
\State \Return \(\left(\hat y_i,\{R_i^{(r)}\}_{r=1}^{3}\right)\)
\end{algorithmic}
\end{algorithm}

%% file: sections/physav_bench.tex
\section{\bench{}}
\label{sec:bench}

\bench{} is designed to evaluate AIGC detectors on event-centric audio-video clips for which physical-law consistency provides a relevant forensic signal.
For example, trajectory-continuity constraints concern how positions and
velocities evolve; energy-decay checks assess rebound trends under dissipative
conditions; and impact-synchronization checks require visible contact and acoustic onset to
align temporally. This benchmark foregrounds event-centric physical behavior rather than face identity or artifacts tied to a single generator.

\begin{figure*}[!t]
  \centering
  \begin{minipage}[c]{0.39\textwidth}
    \centering
    \includegraphics[width=\linewidth]{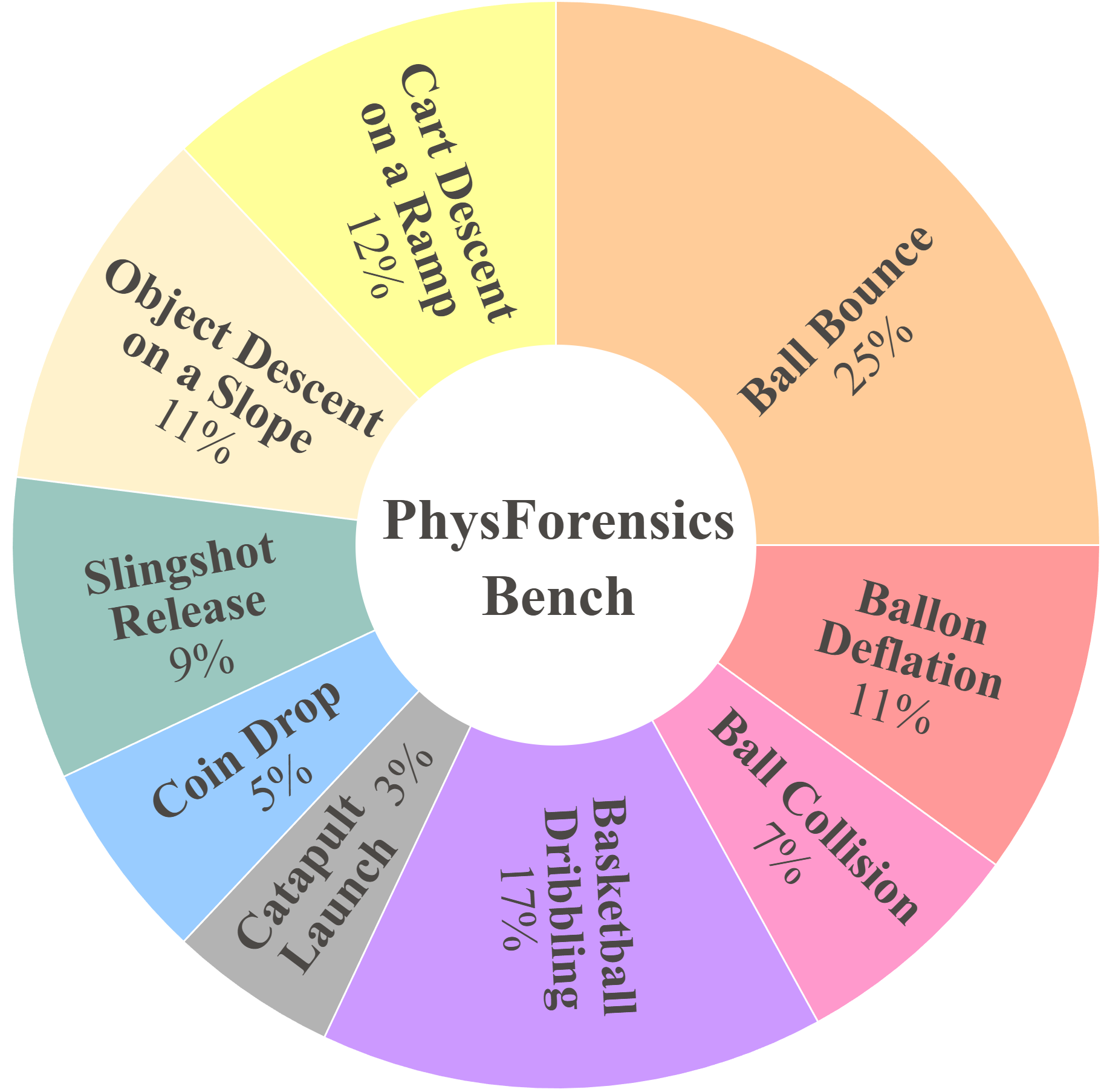}
  \end{minipage}\hfill
  \begin{minipage}[c]{0.59\textwidth}
    \centering
    \includegraphics[width=\linewidth]{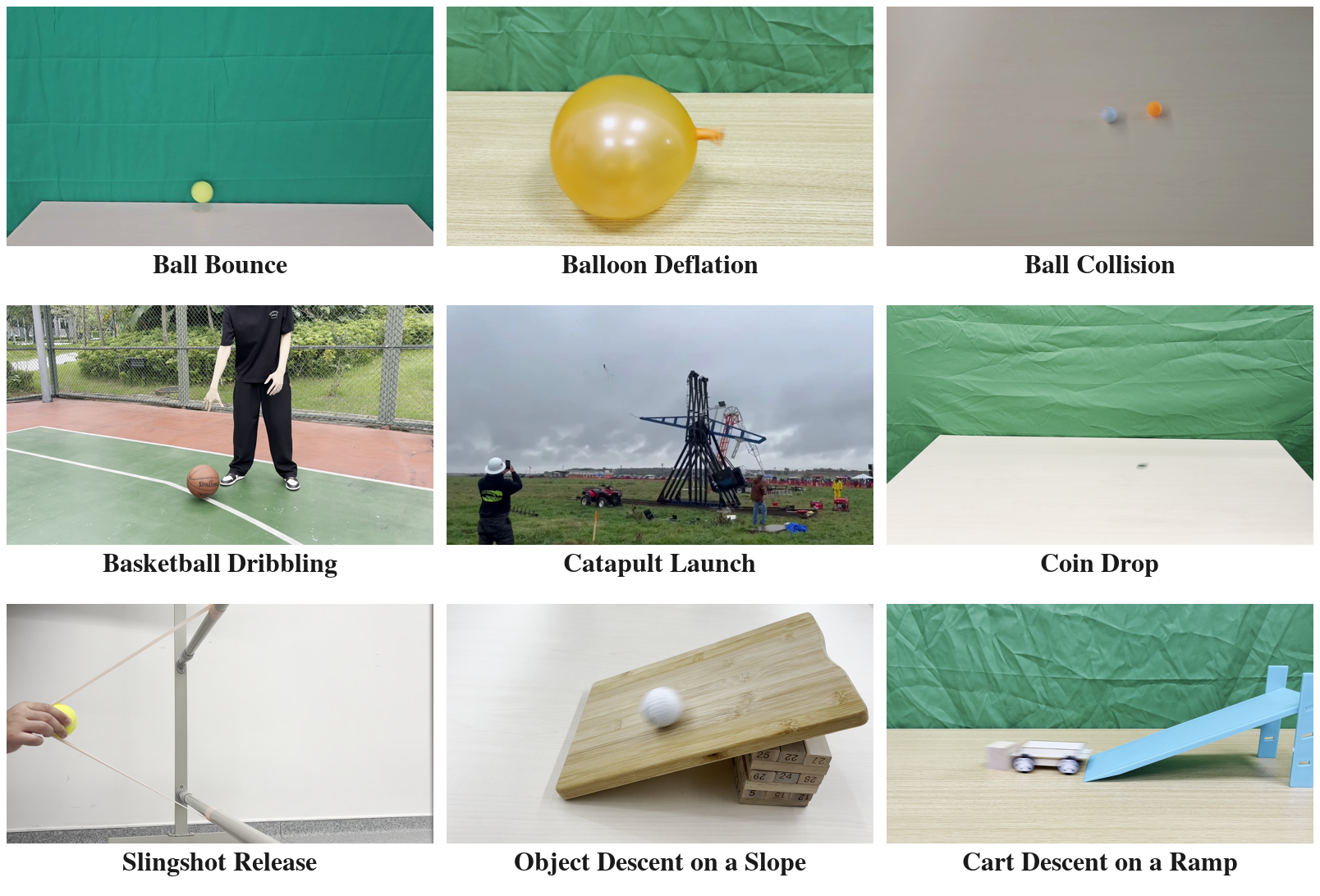}
  \end{minipage}
  \caption{\bench{} overview. Left: a schematic taxonomy of the nine physical
  scene families, where wedge area does not indicate sample count. Right: one
  representative real frame from each scene.}
  \label{fig:scene_grid}
\end{figure*}

\subsection{Data Sources and Generation}

The benchmark contains real captured clips and generated audio-video clips from
Veo~3.1 Fast ~\cite{veo2026} and Seedance 2.0 ~\cite{seedance2026seedance}. For each real clip, we use the first frame as the
visual reference image and provide a text prompt that describes the target
physical event. The generation models then synthesize the full audio-video
continuation.
This design makes each generated clip a counterpart of a real clip: it shares
the same initial visual context and high-level event description, while the
model must synthesize the subsequent visual dynamics and sounds.

The construction is pair-oriented but not perfectly balanced. Some generation
requests were blocked by platform safety or community-policy filters, for
example when the reference frame contained a visible face or when the prompt was
rejected by the platform. As a result, a small number of real clips do not have
all corresponding generated variants. 

\subsection{Physical Scene Taxonomy}

\bench{} covers nine scenes: ball bouncing, balloon deflation, ball collisions,
basketball dribbling, catapult launches, coin drops, slingshot releases,
object descent on a slope, and cart descent on a ramp. The scenes are deliberately
short and event-centric. They create contacts, releases, rebounds, direction
reversals, surface deformation, rolling/sliding acceleration, and repeated
periodic impacts. These events stress different aspects of physical laws:
timing, object relations, kinematic change, material-dependent sound, and
energy trends. Table~\ref{tab:scene_taxonomy} summarizes the scene definitions,
within-scene variation, and representative physical constraints, while
Figure~\ref{fig:scene_grid} provides a visual overview of the nine scene families.
Most scenes are recorded around controlled short physical events. Catapult
Launch is the main exception: its real clips are web-collected from YouTube,
TikTok, and Reddit to cover diverse catapult designs, launch mechanisms, and
viewpoints.

The scene set is designed to separate different kinds of physical quantities.
Ball-bounce and basketball scenes emphasize repeated impact timing and decay;
balloon deflation emphasizes visible deformation and air-release sound;
ball-collision and coin-drop scenes require sharper contact localization; and
slope and ramp scenes require continuous motion reasoning rather than isolated
impact detection. This organization makes the benchmark more than a collection
of object categories: it asks whether detectors can handle several recurring
physical constraints under different objects, surfaces, viewpoints, and motion
patterns.

\input{tables/scene_taxonomy}

\subsection{Benchmark Split}

We split the benchmark at the seed-clip level. Each real seed clip and all of
its available Seedance and Veo counterparts are treated as one group and assigned
to the same train, validation, or test split. This prevents clips that share the
same initial frame and event description from appearing across different splits.
The resulting split contains 878 training samples, 293 validation samples, and
293 test samples, with scene- and source-level distributions reported in
Table~\ref{tab:dataset_stats}. The training split is used
for supervised baseline training, while the validation split supports external
methods that require checkpoint selection or operating-point tuning. \method{}
uses neither split for detector adaptation.
We preserve each clip's native container, codec, frame rate, resolution, duration,
and audio encoding rather than globally re-encoding the benchmark. These
attributes are not supplied to \method{} as detection features: the visual path
operates on decoded frames, infers the source frame rate, and expresses motion on
a millisecond time axis; the audio path operates on decoded mono waveforms,
normalizes amplitude, applies model-specific resampling for SED, and analyzes
high-frequency transients at the native sample rate. File names, source labels,
container metadata, codec identifiers, and bitrates are not included in the
physical quantities or prompts. The generated clips contain no visible
platform watermarks. This processing reduces direct dependence on acquisition
and encoding settings, although we do not assume complete invariance to changes
in compression quality or spatial resolution.

\input{tables/dataset_stats}

%% file: tables/scene_taxonomy.tex
\begin{table*}[!t]
\centering
\caption{Physical scenes, representative physical constraints, and illustrative
relations among physical quantities in \bench{}. The formulas summarize expected
relationships among those quantities rather than a fixed hand-coded law bank; the
Physical Law Selector instantiates measurable constraints for the depicted event at runtime.}
\label{tab:scene_taxonomy}
\footnotesize
\renewcommand{\arraystretch}{1.16}
\setlength{\tabcolsep}{4pt}
\rowcolors{2}{black!4}{white}
\begin{tabular}{@{}>{\centering\arraybackslash}m{0.13\textwidth}
                >{\raggedright\arraybackslash}m{0.29\textwidth}
                >{\raggedright\arraybackslash}m{0.25\textwidth}
                >{\raggedright\arraybackslash}m{0.25\textwidth}@{}}
\toprule
\rowcolor{black!10}
\textbf{Scene} & \textbf{Description} & \textbf{Representative physical constraints} &
\textbf{Illustrative relations} \\
\midrule
Ball Bounce & Ping-pong, tennis, golf, and bouncy balls bounce on a tabletop,
entering from the left, top, or right. &
(1) Impact--sound synchronization; \newline
(2) rebound kinematics; \newline
(3) mechanical and acoustic energy decay. &
(1) \(\min_m |t^a_m-t^v_n|\leq\tau_{\mathrm{sync}}\); \newline
(2) \(v_z(t_n^-)v_z(t_n^+)<0\); \newline
(3) \(H_{n+1}\leq H_n,\ L_{n+1}\leq L_n\). \\

Balloon Deflation & A balloon deflates on a tabletop, either upright or lying
horizontally. &
(1) Deformation--airflow coupling; \newline
(2) deflation-onset synchronization; \newline
(3) airflow and rubber-timbre consistency. &
(1) \(-\dot A(t)\!\uparrow\Rightarrow E_{\mathrm{air}}(t)\!\uparrow\); \newline
(2) \(\min_m|t^a_m-t^v_{\mathrm{defl}}|\leq\tau_{\mathrm{sync}}\); \newline
(3) \(E_{\mathrm{HF}}/E_{\mathrm{tot}}\in\mathcal R_{\mathrm{air/rubber}}\). \\

Ball Collision & An incoming ball is launched from the left or right and
collides with another ball. &
(1) Collision--sound synchronization; \newline
(2) momentum-transfer consistency; \newline
(3) single-collision--single-transient correspondence. &
(1) \(\min_m|t^a_m-t^v_{\mathrm{col}}|\leq\tau_{\mathrm{sync}}\); \newline
(2) \(m_1\Delta\mathbf v_1\approx-m_2\Delta\mathbf v_2\); \newline
(3) \(N_a([t^v_{\mathrm{col}}-\tau,t^v_{\mathrm{col}}+\tau])=1\). \\

Basketball Dribbling & A person dribbles a basketball while stationary,
approaching or leaving the camera, or moving left or right. &
(1) Impact--sound synchronization; \newline
(2) dribble periodicity; \newline
(3) rebound-energy--loudness consistency. &
(1) \(\min_m|t^a_m-t^v_n|\leq\tau_{\mathrm{sync}}\); \newline
(2) \(\operatorname{Var}(\Delta t_n)\leq\epsilon_p\); \newline
(3) \(\operatorname{corr}(H_n,L_n)>0\). \\

Catapult Launch & Web-collected videos show diverse catapult designs and
viewpoints launching a projectile. &
(1) Release synchronization; \newline
(2) projectile-trajectory continuity; \newline
(3) free-flight source and silence consistency. &
(1) \(\min_m|t^a_m-t^v_{\mathrm{rel}}|\leq\tau_{\mathrm{sync}}\); \newline
(2) \(\ddot{\mathbf p}(t)\approx\mathbf g\); \newline
(3) \(N_{\mathrm{unexplained}}(W_{\mathrm{flight}})=0\). \\

Coin Drop & Coins are dropped onto a tabletop or a foam board. &
(1) Impact--sound synchronization; \newline
(2) bounce damping and acoustic decay; \newline
(3) surface-dependent timbre consistency. &
(1) \(\min_m|t^a_m-t^v_n|\leq\tau_{\mathrm{sync}}\); \newline
(2) \(H_{n+1}<H_n,\ L_{n+1}<L_n\); \newline
(3) \(f_c^{\mathrm{table}}>f_c^{\mathrm{foam}}\). \\

Slingshot Release & A slingshot is pulled and released to shoot a ball. &
(1) Release-onset synchronization; \newline
(2) elastic-energy and recoil consistency; \newline
(3) post-release vibration and flight continuity. &
(1) \(\min_m|t^a_m-t^v_{\mathrm{rel}}|\leq\tau_{\mathrm{sync}}\); \newline
(2) \(\tfrac12kx^2\approx\tfrac12mv_0^2+E_{\mathrm{loss}}\); \newline
(3) \(\ddot{\mathbf p}(t)\approx\mathbf g,\ \dot A_{\mathrm{vib}}(t)<0\). \\

Object Descent on a Slope & A ball or wooden block descends a low or high
slope. &
(1) Gravity-driven acceleration; \newline
(2) rolling/sliding friction consistency; \newline
(3) persistent surface-contact continuity. &
(1) \(0<a_{\parallel}\leq g\sin\theta\); \newline
(2) \(|v-r\omega|\leq\epsilon_r\) for rolling; \newline
(3) \(d_{o,s}(t)\leq\epsilon_c\). \\

Cart Descent on a Ramp & A cart descends a ramp, with or without a terminal
wooden block that creates an end collision. &
(1) Gravity-driven descent; \newline
(2) rolling-speed--sound coupling; \newline
(3) terminal-impact synchronization. &
(1) \(0<a_{\parallel}\leq g\sin\theta\); \newline
(2) \(\lambda_{\mathrm{roll}}(t)\propto v(t)/r\); \newline
(3) \(\min_m|t^a_m-t^v_{\mathrm{term}}|\leq\tau_{\mathrm{sync}}\). \\
\bottomrule
\end{tabular}

\vspace{0.35em}
\parbox{0.98\textwidth}{\footnotesize
Here \(t^v\) and \(t^a\) denote visual-event and audio-onset times;
\(H_n\), \(L_n\), and \(\Delta t_n\) denote rebound height, impact loudness,
and inter-impact interval; \(N_a(W)\) counts audio onsets in window \(W\); and
\(\mathcal R\), \(\epsilon\), and \(\tau\) denote applicability-dependent
plausible ranges or tolerances.}
\end{table*}

%% file: tables/dataset_stats.tex
\begin{table}[!t]
\centering
\caption{\bench{} data distribution by scene, split, and source. VEO denotes VEO 3.1 Fast, and Seed. denotes Seedance 2.}
\label{tab:dataset_stats}
\scriptsize
\setlength{\tabcolsep}{2.5pt}
\resizebox{\columnwidth}{!}{%
\begin{tabular}{@{}l r|rrr|rrr@{}}
\toprule
Scene & Total & Train & Val. & Test & Real & VEO & Seed. \\
\midrule
Ball Bounce & 366 & 219 & 72 & 75 & 122 & 122 & 122 \\
Balloon Deflation & 153 & 92 & 31 & 30 & 51 & 51 & 51 \\
Ball Collision & 108 & 66 & 21 & 21 & 36 & 36 & 36 \\
Basketball Dribbling & 242 & 145 & 49 & 48 & 104 & 52 & 86 \\
Catapult Launch & 42 & 24 & 9 & 9 & 14 & 14 & 14 \\
Coin Drop & 74 & 45 & 15 & 14 & 26 & 22 & 26 \\
Slingshot Release & 135 & 81 & 27 & 27 & 45 & 45 & 45 \\
Object Descent on a Slope & 167 & 101 & 33 & 33 & 56 & 55 & 56 \\
Cart Descent on a Ramp & 177 & 105 & 36 & 36 & 59 & 59 & 59 \\
\midrule
Total & 1,464 & 878 & 293 & 293 & 513 & 456 & 495 \\
\bottomrule
\end{tabular}
}
\end{table}

%% file: sections/experiments.tex
\section{Experiments}
\label{sec:experiments}

\subsection{Experimental Setup}

We evaluate on our train/validation/test split described in
Section~\ref{sec:bench}. We report accuracy and F1 score, treating AI-generated
audio-video as the positive class. When a table reports generator-specific
columns, Real+Seedance and Real+Veo denote binary subtasks formed by pairing all
real clips with one generated source.

The main comparison includes methods without \bench{}-specific detector training, \textit{i.e.}, no data from \bench{} is adopted to train or fine-tune the detector parameters. Released checkpoints may nevertheless have been trained on their
original source domains. \method{} uses no \bench{} labels for adaptation.

Unless otherwise stated, \method{} uses the complete video and audio physical
estimator architecture shown in Figure~\ref{fig:physical-estimators}.
Clip captions are produced independently by Gemini 3.1
Pro~\cite{google2026gemini31pro} and GPT-5.5~\cite{openai2026gpt55}, and both
captions are provided to the Physical Law Selector to reduce reliance on a
single model. GPT-5.5 is used for physics-aware sampling strategy selection,
physical-law selection, and physical-consistency verification.
Each sample is verified independently three times to reduce stochastic verifier
failures, and its final Real/Fake prediction is determined by majority voting
over the three run-level labels; confidence values do not weight the vote.
For visual trajectory reconstruction, we set the velocity and acceleration EMA
coefficients to \(\alpha_v=0.3980\) and \(\alpha_a=0.3305\), respectively. Both
were selected on the Unity simulation data without using any
\bench{} labels.

All GPU-based runs, including baseline feature extraction, finetuning, and
physical-estimator preprocessing, are run on eight NVIDIA RTX 3090 GPUs.

\subsection{Main Results}

\input{tables/main_results}

Table~\ref{tab:main_results} reports methods without \bench{} adaptation on the
test split. D3~\cite{zheng2025d3} is a training-free score method, so we use a
validation-real target-FPR threshold to report Accuracy and F1. The direct LLM
rows test prompt-only inspection: GPT-5.5 uses sampled video frames, while
Gemini 3.1 Pro uses video input with audio. \method{} uses no \bench{} labels or
benchmark-specific adaptation.

We report Real+Seedance and Real+Veo subtasks rather than a single overall score
so the table exposes generator-specific behavior and prediction bias. A method
that over-predicts generated videos can have high generated recall but will lose
Accuracy and F1 once real samples are included in the paired subtask. We include
FGI~\cite{astrid2024detecting} to cover the audio-video detector modality; it is a
human-centric audio-video deepfake detector rather than a general AIGC physical
reasoning method, so its result should be read as an out-of-domain reference.

\method{} achieves strong performance on \bench{} without using benchmark labels for detector training or operating-point calibration.
Specifically, \method{} performs best in
all four generator-specific metrics, achieving 70.30\% Accuracy and
64.29\% F1 on Seedance and 72.16\% Accuracy and 65.82\% F1 on Veo.
Gemini 3.1 Pro has relatively high F1 but lower Accuracy, indicating a stronger
bias toward predicting generated videos. Released video detectors and the
out-of-domain FGI detector transfer poorly, showing that source-specific visual
signals or face-centric audio-video cues do not directly solve our
physical-event setting.

\subsection{Per-Scene Analysis}

Table~\ref{tab:per_scene} breaks down \method{} performance by physical scene.
Because \method{} uses no \bench{} labels for adaptation, this diagnostic view
pools all completed training, validation, and test samples to provide more reliable
scene-level estimates. Its values are therefore not directly comparable to the
test-only results in Table~\ref{tab:main_results}.

\input{tables/per_scene_results}

Performance is broadly associated with observability of the underlying physical process. Object descent on a slope is the strongest and most consistent scene, achieving 78.57\%/72.73\% Accuracy/F1 on Seedance and 80.18\%/75.00\% on Veo. Its
sustained trajectory exposes position, velocity, acceleration, and
object-surface relations over many frames. Ball bounce is also comparatively
stable because repeated contacts permit several checks of rebound dynamics and
impact-onset alignment. In contrast, balloon deflation remains below 50\%
Accuracy for both generators: gradual deformation and sustained airflow provide
no sharply localized event, and the relevant shape changes are sensitive to
mask and geometry estimation.

Coin drop reveals a different, generator-specific failure mode. The two paired
subtasks both correctly classify 21 of 26 real clips, so their gap comes
entirely from generated clips: \method{} detects only 3 of 26 Seedance clips as
fake, compared with 13 of 22 Veo clips. The Veo reports frequently identify a
major audio onset before or after the visible surface contact, a missing onset at
the first impact, or a late strong transient after the coin has nearly stopped.
Seedance clips more often exhibit aligned first impacts and plausible acoustic
decay; later sounds can be attributed to rolling or friction, while weak
unmatched events are usually marked uncertain rather than violated. This yields
17.65\% F1 on Seedance but 65.00\% on Veo.

Slingshot release shows the same diagnosis: the real-clip count is again fixed
at 42 of 45, whereas generated-clip detections increase from 10 of 45 for
Seedance to 28 of 45 for Veo. Thus, the generator gaps do not indicate a general
bias on real videos; they measure how often each generator exposes violations
that are observable through the selected quantities and laws. Overall,
verification is strongest for sustained trajectories and repeated events, but
weaker for subtle deformation, small objects, and brief contacts. These results
motivate more accurate physical-quantity estimation for weakly observed events
and should not be interpreted as a generator-wide quality ranking.

%% file: tables/main_results.tex
\begin{table}[!t]
\centering
\scriptsize
\caption{Main test-set results for methods without \bench{} adaptation. Seedance and VEO denote the Real+Seedance and Real+VEO binary subtasks, respectively. Accuracy and F1 are reported as percentages.}
\label{tab:main_results}
\setlength{\tabcolsep}{3pt}
\begin{tabularx}{\columnwidth}{@{}llYY|YY@{}}
\toprule
Method & Modality & \multicolumn{2}{c}{Seedance} & \multicolumn{2}{c}{VEO} \tabularnewline
\cmidrule(lr){3-4}\cmidrule(lr){5-6}
 & & Acc(\%) & F1(\%) & Acc(\%) & F1(\%) \tabularnewline
\midrule
NSG-VD~\cite{zhang2026physics} & Video & 48.02 & 21.05 & 52.06 & 27.91 \tabularnewline
D3~\cite{zheng2025d3} & Video & \underline{65.35} & 52.05 & \underline{65.46} & 49.62 \tabularnewline
FGI~\cite{astrid2024detecting} & Audio-video & 49.50 & 22.73 & 45.88 & 7.08 \tabularnewline
GPT-5.5~\cite{openai2026gpt55} & Video & 51.49 & 2.00 & 55.67 & 10.42 \tabularnewline
Gemini 3.1 Pro~\cite{google2026gemini31pro} & Audio-video & 53.96 & \underline{60.09} & 57.22 & \underline{63.44} \tabularnewline
\method{} & Audio-video & \textbf{70.30} & \textbf{64.29} & \textbf{72.16} & \textbf{65.82} \tabularnewline
\bottomrule
\end{tabularx}
\end{table}

%% file: tables/per_scene_results.tex
\begin{table}[!t]
\centering
\scriptsize
\caption{Per-scene diagnostic results for \method{} pooled over all completed train, validation, and test samples. Accuracy and F1 are reported as percentages.}
\label{tab:per_scene}
\setlength{\tabcolsep}{4pt}
\begin{tabularx}{\columnwidth}{@{}lYY|YY@{}}
\toprule
Scene & \multicolumn{2}{c}{Seedance} & \multicolumn{2}{c}{VEO} \tabularnewline
\cmidrule(lr){2-3}\cmidrule(lr){4-5}
 & Acc(\%) & F1(\%) & Acc(\%) & F1(\%) \tabularnewline
\midrule
Ball Bounce & 68.03 & 69.29 & 63.52 & 63.37 \tabularnewline
Balloon Deflation & 48.04 & 51.38 & 45.10 & 47.17 \tabularnewline
Ball Collision & 69.44 & 62.07 & 68.06 & 59.65 \tabularnewline
Basketball Dribbling & 67.37 & 62.20 & 77.56 & 71.54 \tabularnewline
Catapult Launch & 64.29 & 61.54 & 71.43 & 71.43 \tabularnewline
Coin Drop & 46.15 & 17.65 & 70.83 & 65.00 \tabularnewline
Slingshot Release & 57.78 & 34.48 & 77.78 & 73.68 \tabularnewline
Object Descent on a Slope & 78.57 & 72.73 & 80.18 & 75.00 \tabularnewline
Cart Descent on a Ramp & 70.34 & 57.83 & 70.34 & 57.83 \tabularnewline
\bottomrule
\end{tabularx}
\end{table}

%% file: sections/limitations.tex
\section{Future Directions} 

Our results indicate that physics-grounded verification is a promising direction for AIGC detection. The structured quantity representation in \method{}, the selection of physical laws based jointly on event relevance and measurability, and the explicit verification records returned by the framework together provide a foundation for several future directions. 

\noindent\textbf{Broader physical coverage.} The current implementation estimates object extent, three-dimensional kinematics, and event-level acoustic quantities, enabling the verification of constraints concerning event-onset anchoring, kinematic and energy continuity, material-acoustic consistency, source plausibility, and scene geometry. Future work can extend this foundation with direct estimators for physical quantities and properties that are not yet explicitly recovered, such as mass, volume, density, material properties, elasticity parameters, frictional parameters, and contact force. Additional quantity fields could support constraints involving illumination, reflection, refraction, acoustic propagation, resonance, and observable electromechanical processes. As these quantities become reliably measurable, the same selection-and-verification framework can expand from its current mechanics- and acoustics-oriented coverage to broader families of physical laws. 

\noindent\textbf{Uncertainty-aware and executable verification.} A second direction is to make verification explicitly aware of measurement uncertainty. Future estimators can report calibrated confidence intervals, observation quality, and alternative measurement hypotheses together with their estimated quantities. The verifier can then propagate this information when assigning supported, violated, or uncertain states. Disagreement among estimators can expose measurement ambiguity, whereas consistent violations across independently estimated quantities or related constraints can strengthen the evidence of physical inconsistency. For constraints that admit precise numerical forms, semantic reasoning can also be separated from numerical execution: reasoning models can determine which laws are relevant to the depicted event and verifiable from the available quantities, while unit-aware executable operators evaluate equations, inequalities, trends, and temporal relations reproducibly. This combination would retain the flexibility of event-aware, measurability-constrained law selection while making applicable numerical checks more reproducible and inspectable. 

\noindent\textbf{Broader evaluation and longitudinal testing.} Future versions of \bench{} can expand the physical domains, object categories, capture conditions, acoustic environments, and generator families represented in the benchmark. Matched cross-generator, cross-scene, and cross-law-family protocols can directly evaluate whether the verification framework remains effective on previously unseen generators and events. Longitudinal evaluation across successive generations of models can further examine whether the selected law families, measurable quantities, and observed violation patterns remain informative as generation technologies evolve. Such evaluations would directly test the robustness to evolving generators that motivates physics-grounded AIGC detection.

%% file: sections/conclusion.tex
\section{Conclusion} 

We presented \method{}, a physics-grounded verification framework for AIGC detection that tests whether the depicted event is consistent with measurable constraints derived from physical laws. \method{} estimates structured physical quantities from the video and audio streams, selects laws that are relevant to the event and verifiable from the available quantities, and assesses each applicable constraint as supported, violated, or uncertain. Beyond a real/fake decision, it returns supporting evidence specifying the tested law, relevant time window, verification outcome, and quantities used in the assessment.

We also introduced \bench{}, comprising paired real and generated audio-video clips across nine event-centric scene families, with generated samples produced by Seedance~2.0 and Veo~3.1 Fast. On this benchmark, \method{} achieves the highest accuracy and F1 score among the evaluated methods in the primary comparison, outperforming direct foundation-model inspection and the compared baselines. These results provide empirical support for physical-consistency verification as a practical strategy for AIGC detection.

More broadly, this work establishes event-aware, measurability-constrained physical-consistency verification as a complementary direction for AIGC detection. It connects physical perception with multimedia forensics and provides a framework in which advances in physical-quantity estimation and physical-law coverage can be incorporated as structured, inspectable evidence. This perspective opens a promising research direction for building AIGC detectors around explicit evidence from the physical behavior of depicted events, which may remain informative as generated media become increasingly realistic.

%% file: sections/appendix.tex
\label{app:process}
\raggedbottom
\suppressfloats[t]

\subsection{LLM Agent Workflow}

\method{} uses prompted models inside the agent flow described in
Section~\ref{sec:method}. Captioning first generates clip captions
from ordered frames or video input. Physical Quantity Estimation converts the
raw streams into video and audio quantities. The Physical Law Selector then
uses the clip captions and available quantity fields to select only laws whose constraints can be evaluated from those quantities. The Physical Consistency Verifier
receives the selected laws and compressed quantities, checks the corresponding
constraints, and reports support, violations, uncertainty, and a final real/fake verdict.

\subsection{Notation and Implementation Mapping}
\label{app:notation}
We summarize the symbols used in Section~\ref{sec:method} before presenting the
prompt examples. Prompted model mappings use calligraphic letters with
\(\theta\), which denotes the model and prompting configuration of the
corresponding stage rather than a shared trainable parameter tensor. Vector
quantities use boldface, scalar quantities use ordinary italic letters, and
deterministic procedures use upright function names.

The concrete backends used in our experiments are listed in
Table~\ref{tab:model_notation}, while Table~\ref{tab:symbol_notation} defines
the remaining symbols.

\begin{table}[t]
\caption{Prompted model mappings and their concrete implementations.}
\label{tab:model_notation}
\centering
\scriptsize
\renewcommand{\arraystretch}{1.20}
\setlength{\tabcolsep}{3pt}
\rowcolors{2}{black!4}{white}
\begin{tabular}{>{\centering\arraybackslash}m{0.12\columnwidth}
                >{\centering\arraybackslash}m{0.24\columnwidth}
                >{\raggedright\arraybackslash}m{0.55\columnwidth}}
\toprule
\rowcolor{black!10}
\textbf{Symbol} & \textbf{Mapping} & \textbf{Implementation and role} \\
\midrule
\(\mathcal{B}_{\theta}\) & Captioning mapping & Gemini 3.1 Pro and GPT-5.5 independently describe the clip; \(B_i\) retains both captions for downstream physical-law selection. \tabularnewline
\(\mathcal{P}_{\theta}\) & Physical Law Selector & GPT-5.5 selects an event-relevant and measurable law set \(L_i\) and encodes each physical constraint in its \texttt{decision\_basis}. \tabularnewline
\(\mathcal{V}_{\theta}\) & Physical Consistency Verification and report generation & GPT-5.5 evaluates the selected laws and constraints and produces the run-level label, verification records, and summary in one API call. Each clip is evaluated in three independent calls before local majority voting. \tabularnewline
\bottomrule
\end{tabular}

\end{table}

\begin{table}[t]
\caption{Notation used throughout the method. Section rows group symbols by their role in the pipeline.}
\label{tab:symbol_notation}
\centering
\scriptsize
\renewcommand{\arraystretch}{1.20}
\setlength{\tabcolsep}{3pt}
\rowcolors{2}{black!4}{white}
\begin{tabular}{>{\centering\arraybackslash}m{0.43\columnwidth}
                >{\raggedright\arraybackslash}m{0.51\columnwidth}}
\toprule
\rowcolor{black!10}
\textbf{Symbol} & \textbf{Definition and role} \\
\midrule
\rowcolor{black!8}
\multicolumn{2}{l}{\textbf{Core inference notation}} \tabularnewline
\(i,r\) & Clip index and verification-run index, respectively. \tabularnewline
\(\mathbf{x}_i\) & Input audio-video clip. \tabularnewline
\(B_i\) & Captions retained for clip \(i\). \tabularnewline
\(M_i=(A_i,V_i)\) & Full event-level audio and frame-level video quantities. \tabularnewline
\(\tilde M_i\) & Compact physical quantities supplied to law selection and verification. \tabularnewline
\(L_i\) & Physical laws and constraints selected for clip \(i\). \tabularnewline
\(\hat y_i^{(r)},R_i^{(r)}\) & Binary label and structured report from verification run \(r\). \tabularnewline
\(\hat y_i,R_i\) & Majority-voted label and the retained reports from all three runs. \tabularnewline
\midrule
\rowcolor{black!8}
\multicolumn{2}{l}{\textbf{Video-estimator implementation notation}} \tabularnewline
\(o,(u,v),t\) & Object index, image-plane pixel coordinates, and frame time. \tabularnewline
\(\Omega_{o,t}\) & Object-mask pixels at time \(t\). \tabularnewline
\(D_t,K_t,\hat T_t\) & Metric depth map, \(3\times3\) camera-intrinsic matrix, and \(4\times4\) world-to-camera transform. \tabularnewline
\(\mathbf p_t^c,\mathbf p_t^w\) & Reconstructed 3D point in the camera frame and the fixed world frame. \tabularnewline
\(\bar{\mathbf u}_{o,t},\bar D_{o,t},\mathbf c_{o,t}\) & Mask centroid, median metric depth, and reconstructed object center. \tabularnewline
\(\mathbf v^{\mathrm{raw}}_{o,t},\mathbf a^{\mathrm{raw}}_{o,t}\) & Raw velocity and acceleration from backward finite differences. \tabularnewline
\(\mathbf v_{o,t},\mathbf a_{o,t}\) & EMA-smoothed object velocity and acceleration. \tabularnewline
\(\Delta t,\alpha_v,\alpha_a\) & Frame-time step and velocity/acceleration EMA coefficients. \tabularnewline

\bottomrule
\end{tabular}
\end{table}

\subsection{Prompt and Response Examples}
\label{app:prompt-examples}

This section reports representative prompt structures used by the main LLM
agents and one verifier response example. Runtime prompts are filled with video
frames, clip captions, sanitized audio/video physical quantities, and selected
laws. For readability, the templates are translated
and lightly abridged while preserving their task instructions, constraints, and
output schemas.

\Needspace{5\baselineskip}
\Needspace{8\baselineskip}
\exampleheading{Captioning prompt.}
The captioning prompt asks each model to produce a factual description of the
clip.
\begin{lstlisting}[style=promptbox]
Review the provided clip and produce a factual clip caption.
Output JSON only:
{
  "caption": string
}
Constraints:
- The caption should be a concise factual paragraph, with no Markdown.
- Describe the environment, objects, actions, interactions, and camera/view changes.
- Follow the video timeline from beginning to end and include key transitions.
- Keep it factual and do not introduce unsupported details.
- Use the requested output language.
\end{lstlisting}
\par\medskip

\Needspace{8\baselineskip}
\exampleheading{Physical-quantity input examples.}
The verifier receives compressed physical quantities rather than raw
waveforms, raw frames, or detector logits. The following abridged snippets come
from the same generated ball-bounce example used below. They illustrate the
detailed records exposed to physical consistency verification; the Physical Law
Selector receives only a compact summary of their available fields.

\emph{Audio event JSON} keeps the aligned time origin, onset, acoustic
envelope, detector sources, and native-rate spectral descriptors.
\begin{lstlisting}[style=jsonbox]
{
  "audio_meta": {
    "sample_rate": 48000,
    "duration_ms": 4020,
    "time_origin": "aligned_with_video_frame_0"
  },
  "events": [
    {
      "event_id": 0,
      "onset_ms": 20,
      "active_start_ms": 0,
      "active_end_ms": 363,
      "duration_ms": 363,
      "peak_db": -23.566,
      "features": {
        "rms_mean": 0.014,
        "hnr_mean": -31.361,
        "centroid_mean": 3963.245,
        "decay_rate_ms": 93,
        "spectral_band": "0-8000Hz",
        "high_band_delta_db": 60.378,
        "spectral_novelty_score": 60.378,
        "sample_rate_used": 48000
      },
      "labels": {
        "sed_primary_label": "Knock",
        "event_type": "impact",
        "detector_sources": ["pretrained_sed", "spectral_proposal"]
      }
    },
    ...
  ]
}
\end{lstlisting}
\par\medskip

\Needspace{24\baselineskip}
\emph{Compressed video-quantity CSV} provides object identities, sampled
trajectories, and relation distances. The example shows a ball and table; small
\texttt{min\_dist} values around audio onsets support impact-synchronization
checks, while object velocity and acceleration support trajectory-continuity
checks. This is an excerpt from the sparse visual quantity table used by the
verifier.
\begin{lstlisting}[style=promptbox,breakatwhitespace=false,literate={,}{{,}\allowbreak}1]
[Object Names]
obj_id,obj_name
1,<ball-1>
2,<table-2>

[Object Trajectories]
t_ms,obj_id,x,y,z,vx,vy,vz,ax,ay,az,distance_to_camera,mask_area_px
...
458,1,-0.829,0.152,1.301,0.470,1.850,-0.448,3.374,2.698,-5.235,1.551,13630
458,2,-0.038,0.365,1.128,-0.001,-0.004,-0.007,-0.006,0.076,0.137,1.188,420179
500,1,-0.796,0.229,1.256,0.597,1.845,-0.695,2.881,0.043,-5.465,1.508,12233
500,2,-0.038,0.365,1.126,0.001,-0.007,-0.022,0.035,-0.059,-0.309,1.187,418449
542,1,-0.761,0.302,1.226,0.759,1.720,-0.712,3.337,-2.767,-3.471,1.465,14292
...

[Relative Distances]
t_ms,obj1_id,obj2_id,centroid_distance,min_dist
...
458,1,2,0.837,0.093
500,1,2,0.781,0.011
542,1,2,0.723,0.004
625,1,2,0.756,0.028
667,1,2,0.776,0.083
...
\end{lstlisting}
\par\medskip

\Needspace{8\baselineskip}
\exampleheading{Physical-law-selection prompt.}
The Physical Law Selector receives the clip captions and a compact field summary
of the available physical quantities. It selects laws before seeing the full
numerical tables, so every \texttt{decision\_basis} constraint must be evaluable
from the available fields.
\begin{lstlisting}[style=promptbox]
You are a physics-grounded audio-video law selector.

Task:
Do not perform verification in this stage. Using only the clip captions and the available audio/video quantity-field summary, select which physical laws should be checked later and assign each law a relative weight. Consider both clip semantics and observable physical quantities. Exclude any law whose required quantities are unavailable.

Selection requirements:
- Select approximately three applicable physical-law checks as a prompting target; the returned law set may have a different size.
- For each law, specify physical basis, audio fields, video fields, applicability, failure conditions, decision basis, and weight.
- Law weights must sum to approximately 1.
- Return JSON only.

Output schema:
{
  "meta": {
    "scene_understanding": "string",
    "selection_principles": ["string"]
  },
  "law_plan": [
    {
      "law_name": "string",
      "display_name": "string",
      "weight": 0.5,
      "physical_basis": "string",
      "audio_fields": ["string"],
      "video_fields": ["string"],
      "decision_basis": "string",
      "applicability": "string",
      "failure_conditions": "string"
    }
  ],
  "selection_summary": "string"
}

Runtime input:
<clip captions and audio/video quantity-field summary>
\end{lstlisting}
The field \texttt{decision\_basis} states the operational constraint;
Figure~\ref{fig:pipeline} shows representative symbolic forms.
\par\medskip

\noindent\begin{minipage}{\columnwidth}
\exampleheading{Physical-consistency-verification prompt.}
The Physical Consistency Verifier receives the selected laws, their
\texttt{decision\_basis} constraints, and compressed quantity tables. It is
instructed to evaluate only those constraints rather than select new laws.
\begin{lstlisting}[style=promptbox]
You are a physics-grounded audio-video law verifier.

The preceding selection stage has already selected laws and weights from the clip captions and the observable fields. You may only use the given law names, physical bases, required audio/video fields, decision bases, applicability, failure conditions, and weights. Do not add new laws.

If a law is irrelevant to a time window, omit it for that window. If it is relevant but evidence is insufficient, output uncertain. Mild deviations should be uncertain or low-severity; only strong evidence should produce a violation.

For every decision, cite the relevant quantity fields, timestamps, and measured values when available.

\end{lstlisting}
\end{minipage}
\par\medskip

\noindent\begin{minipage}{\columnwidth}
\begin{lstlisting}[style=promptbox]
Return JSON only:
{
  "meta": {
    "window_ms": number,
    "audio_duration_ms": number,
    "video_duration_ms": number
  },
  "window_analysis": [
    {
      "window_ms": [number, number],
      "laws": {
        "<law_name>": {
          "decision": "support|violate|uncertain",
          "confidence": number,
          "evidence": "string"
        }
      }
    }
  ],
  "violations": [
    {"time_ms": number, "law": "string", "evidence": "string", "severity": "low|medium|high"}
  ],
  "final_result": {
    "verdict": "physical_plausible|partially_plausible|physically_implausible",
    "real_or_fake": "Real|Fake",
    "summary": "string"
  }
}

Runtime input:
<selected laws, clip captions, field descriptions, audio CSV, video CSV>
\end{lstlisting}
\end{minipage}
\par\medskip

\Needspace{8\baselineskip}
\exampleheading{Verifier output example.}
The following is an abridged example of a final verification response for a
generated ball-bounce clip. The response is translated from the original model
output and shows how \method{} exposes the physical reason behind the decision,
rather than returning only a binary label.
\begin{lstlisting}[style=jsonbox]
{
  "violations": [
    {
      "time_ms": 20,
      "law": "impact_synchronization",
      "evidence": "A strong isolated audio onset appears at the beginning of the clip, but the visible ball has not yet contacted the table or any other surface. No plausible visible source explains the transient.",
      "severity": "high"
    },
    {
      "time_ms": 1330,
      "law": "bounce_energy_dissipation_consistency",
      "evidence": "The second visible impact has lower visual energy than the previous bounce, but the corresponding audio transient becomes stronger. This is inconsistent with a simple dissipative bouncing sequence.",
      "severity": "medium"
    }
  ],
  "final_result": {
    "verdict": "physically_implausible",
    "real_or_fake": "Fake",
    "summary": "Most bounce onsets align with visible table contacts and rebound candidates, and the trajectory is broadly gravity-consistent. However, an early unsourced transient and an energy-ordering inconsistency indicate imperfect audio-video physical plausibility."
  }
}
\end{lstlisting}
\par\medskip

\subsection{Video Physical Quantity Estimator}
\label{app:video-estimator}

The video physical estimator combines a frozen object-understanding path with a
simulation-calibrated geometry path. GroundingDINO detects scene objects from
text labels~\cite{liu2024grounding}, and SAM~2 propagates the resulting masks
bidirectionally through the clip~\cite{ravi2025sam}. Frozen ViPE estimates camera
intrinsics, camera motion, and dense near-metric depth. A lightweight
ridge-regression adapter calibrates predicted intrinsics against known camera
parameters, while a residual convolutional adapter predicts a log-depth
correction from frozen depth and camera-aware pixel coordinates. The mask models
are pretrained, and the intrinsics/depth adapters are fitted or trained only on
the Unity simulation calibration set described below; none is trained on
\bench{} real/fake labels.

\paragraph{Geometric reconstruction and kinematics.}
For reproducibility, we next specify how the concrete video backend converts
its mask and geometry outputs into the kinematic quantities consumed by \method{}.
Let \(\Omega_{o,t}\) be the mask pixels of object \(o\) at frame time \(t\),
\(D_t(u,v)\) the depth map, \(K_t\in\mathbb{R}^{3\times3}\) the
camera-intrinsic matrix, and \(\hat T_t\in
SE(3)\subset\mathbb{R}^{4\times4}\) the world-to-camera transform. Depth
values and reconstructed 3D coordinates are expressed in meters. The camera
frame follows the OpenCV convention, with right, down, and forward as positive
\(x\), \(y\), and \(z\), respectively. ViPE establishes a fixed world frame
from its reference camera. Each masked pixel is back-projected into the camera
frame and then mapped into this world frame:
\begin{equation}
\begin{aligned}
\mathbf p_t^c(u,v)
&= D_t(u,v)K_t^{-1}[u,v,1]^\top
\in \mathbb{R}^3, \\
\mathbf p_t^w(u,v)
&= \left(
\hat T_t^{-1}
\begin{bmatrix}\mathbf p_t^c(u,v)\\1\end{bmatrix}
\right)_{1:3},
\qquad (u,v)\in\Omega_{o,t}.
\end{aligned}
\label{eq:backprojection}
\end{equation}
The appended 1 allows rotation and translation in one matrix multiplication,
while \((\cdot)_{1:3}\) retains the resulting 3D world coordinate. The object
center is estimated from the 2D mask centroid and median object depth:
\begin{equation}
\begin{aligned}
\bar{\mathbf{u}}_{o,t}
&= \frac{1}{|\Omega_{o,t}|}
\sum_{(u,v)\in\Omega_{o,t}}[u,v,1]^\top, \\
\bar D_{o,t}
&= \operatorname{median}
\{D_t(u,v):(u,v)\in\Omega_{o,t}\}, \\
\mathbf{c}_{o,t}
&= \left(
\hat T_t^{-1}
\begin{bmatrix}
\bar D_{o,t}K_t^{-1}\bar{\mathbf u}_{o,t}\\1
\end{bmatrix}
\right)_{1:3}.
\end{aligned}
\label{eq:object-center}
\end{equation}
Raw velocities and accelerations are computed by backward finite differences
and smoothed with separate exponential moving averages to suppress derivative
noise from frame-level depth and mask jitter while retaining the motion trend:
\begin{equation}
\begin{aligned}
\mathbf{v}^{\mathrm{raw}}_{o,t}
&= \frac{\mathbf{c}_{o,t}-\mathbf{c}_{o,t-\Delta t}}{\Delta t}, \\
\mathbf{a}^{\mathrm{raw}}_{o,t}
&= \frac{\mathbf{v}^{\mathrm{raw}}_{o,t}
-\mathbf{v}^{\mathrm{raw}}_{o,t-\Delta t}}{\Delta t}, \\
\mathbf{v}_{o,t}
&= \alpha_v \mathbf{v}^{\mathrm{raw}}_{o,t}
+(1-\alpha_v)\mathbf{v}_{o,t-\Delta t}, \\
\mathbf{a}_{o,t}
&= \alpha_a \mathbf{a}^{\mathrm{raw}}_{o,t}
+(1-\alpha_a)\mathbf{a}_{o,t-\Delta t}.
\end{aligned}
\label{eq:finite-difference}
\end{equation}
Here, \(\alpha_v\) and \(\alpha_a\) control velocity and acceleration
smoothing; their values are reported in Section~\ref{sec:experiments}.

The video ablation removes these enhancement components: it restores ViPE's
default Track-Anything-style mask path~\cite{yang2023track}, where
GroundingDINO-seeded SAM masks are propagated by AOT/DeAOT
~\cite{kirillov2023segment,yang2021associating,yang2022decoupling}, and bypasses
the intrinsics and depth adapters. The downstream physical quantity estimation,
sampling, law-selection, and verification stages remain unchanged.

\subsection{Unity Simulation Calibration Data}
\label{app:simulation-calibration}

The simulation calibration set is used only to improve the quality of video-side
quantity estimates. It is not part of \bench{} and it does not contain real/fake labels.
The source is a Unity ball-dropping simulation archive with known camera
parameters, camera poses, object identities, object masks, and depth. We convert
each simulated episode into a ViPE-style monocular run by exporting RGB frames,
depth, object masks, camera pose, camera intrinsics, and object metadata. Each
episode is rendered from four fixed cameras, so the final calibration set
contains 260 physical episodes and 1,040 monocular object runs.

Splits are assigned by Unity simulation episode rather than by camera view. This
prevents the same physical trajectory from appearing in both training and
validation/test through different cameras. The split contains 208/26/26 episodes
for train/validation/test, corresponding to 832/104/104 monocular runs. The
object labels are common ball-like objects: basketball, baseball, golf ball,
ping-pong ball, soccer ball, tennis ball, and volleyball. Each run contributes
90 frames with ground-truth depth, masks, intrinsics, camera pose, and object
identity.

This supervision supports three video-estimator calibration choices. First, the
GroundingDINO+SAM~2 bidirectional mask path is selected because it produces much
cleaner object masks on held-out simulation runs than the default ViPE mask
path. Second, the intrinsics adapter learns a small ridge-regression correction
from predicted camera parameters to known camera parameters. Third, the residual
depth adapter learns to correct frozen depth predictions using predicted depth
and camera-aware pixel coordinates. These calibrations are applied before video
physical quantity estimation on \bench{} clips, but no \bench{} labels are
used in fitting or selecting them.

In the held-out Unity simulation audit, bidirectional GroundingDINO+SAM~2 masks
achieved about \(0.94\) foreground IoU, while the default masks were about
\(0.54\)--\(0.55\). The residual depth adapter reduced full-frame absolute
relative error from \(0.393\) to \(0.057\) and object-region absolute relative
error from \(0.450\) to \(0.188\). These results characterize physical-quantity estimation rather
than \bench{} detection results; they justify the calibrated components used by
the final video physical estimator.

\subsection{Audio Physical Quantity Estimator}
\label{app:audio-evidence}

The audio JSON uses the same time origin as the video: all timestamps are aligned
to video frame 0. The semantic path uses the PretrainedSED inference framework
with a BEATs backbone~\cite{schmid2025effective,chen2022beats}, following AudioSet
and AudioSet Strong sound-event supervision~\cite{gemmeke2017audio,hershey2021benefit}.
The input waveform is resampled to
16 kHz and processed in 10 s chunks, producing frame-level sound-event
probabilities. We decode these probabilities into coarse semantic event regions
with a probability threshold of \(0.2\), a median filter window of 9 frames, and
a 50 ms minimum event duration. The audio agent then refines these coarse
regions with waveform onsets and envelope boundaries.

The native-rate spectral path is designed for physical sounds that are brief,
weakly semantic, or high-frequency. It computes STFT band novelty on the
original waveform with a 25 ms window and a 10 ms hop. The active configuration
uses bands \(0\)--\(8\) kHz, \(8\)--\(16\) kHz, and \(16\)--\(24\) kHz, 
skipping bands above Nyquist when the source sample rate is
lower. For each band, power is converted to dB and compared with a causal
120 ms median baseline:
\begin{equation}
\Delta_b(t) = P_b(t) -
\operatorname{median}\!\left(P_b(t') : t-120\text{ ms} \le t' < t\right).
\label{eq:spectral-novelty}
\end{equation}
A spectral proposal must pass a robust threshold of median novelty plus \(6\)
MAD, at least \(6\) dB energy rise, and a peak floor of \(-75\) dB. Proposal
regions are padded by 20 ms, filtered to be at least 30 ms long, and merged
within an 80 ms window. The proposal label is the neutral
\path{spectral_transient}; it does not inject a semantic sound class.

The main audio branch merges semantic and spectral proposals before feature
extraction. Two regions are fused when they overlap or when their boundary gap is
within 80 ms. The fused window is the temporal union of the regions; if a
semantic label is present, it remains the primary semantic label, while
\texttt{detector\_sources} records whether the event was proposed by
\texttt{pretrained\_sed}, \texttt{spectral\_proposal}, or both. The event record
keeps detector-source metadata, peak and RMS energy, HNR, spectral centroid,
flux, zero-crossing rate, decay estimate, dominant spectral band, high-band
energy rise, spectral novelty score, and the sample rate used by the spectral path.
Spectral-only events are kept as physical transient candidates with no
fabricated semantic label. Events proposed only by the semantic detector still
receive native-rate spectral descriptors on their final event window, so weak
or absent high-band content is represented by measured values rather than by
deleting the field. The audio ablation removes the native-rate
spectral-transient proposal branch and its spectral-only event fields while
retaining the semantic event detector.

\subsection{Physics-Aware Quantity Compression}
\label{app:physical-aware-sampling}

The full video quantity stream contains frame-level trajectories and relations,
so directly placing every record in the verification prompt is unnecessarily
expensive. Using the clip captions \(B_i\), the sampler selector chooses from a
fixed catalog of policies. The selected policy is then applied deterministically
to \(M_i\) to produce the compact representation \(\tilde M_i\).
The selector receives neither the real/fake label nor source-bearing sample
identifiers. The catalog contains six choices: \emph{relation extrema}, which
retain object-pair distance extrema; \emph{kinematic extrema}, which retain
speed, acceleration, direction-change, and trajectory-turning events;
\emph{kinematic--relation sparse}, which combines the first two;
\emph{area-change extrema}, which retain visible-area and deformation changes;
\emph{temporal uniform}, which provides uniform coverage when no reliable event
cue exists; and \emph{hybrid-event sparse}, which combines relation, kinematic,
and area-change candidates.

The selected deterministic sampler extracts visual core frames and retains the
two neighboring frames on each side. The selected visual quantities are then
converted from JSON into a compact CSV table. Audio quantities require less
aggressive temporal sampling because the audio estimator first detects event
intervals and then computes event-level quantities; these records are likewise
converted from JSON into CSV. The two compact tables, together with sampling
metadata, form \(\tilde M_i\). This metadata tells the verifier that sparse
visual rows are deliberately event-centered rather than missing dense
observations. On a stratified 135-clip sample, compression reduces the
verification input from approximately 154.3k to 8.9k tokens per clip, or about
94\%. Appendix~\ref{app:compute-cost} provides the detailed calculation.

\subsection{Physical Quantity Estimator Ablations}
\label{app:estimator-ablations}

\input{tables/ablation_results}

Table~\ref{tab:ablations} measures the contribution of estimator enhancements by
removing them from the complete architecture. Without video enhancement,
Accuracy/F1 decrease from 70.30\%/64.29\% to 60.40\%/54.55\% on Seedance and
from 72.16\%/65.82\% to 62.89\%/57.14\% on VEO. This ablation restores the
default mask path and removes the Unity-calibrated intrinsics and depth adapters,
showing that calibrated object geometry is important for reliable physical
dynamics. Removing high-band transient detection yields 64.85\%/60.34\% on
Seedance and 65.46\%/59.88\% on VEO. High-band transients therefore complement
semantic sound events with additional physical timing and spectral information
for physical-consistency verification.

\subsection{Qualitative Verification of Physical Constraints}
\label{app:qualitative-constraints}

Each selected law includes a \texttt{decision\_basis} that states its physical
constraint over the required quantities. The formulas in
Figure~\ref{fig:pipeline} and Table~\ref{tab:scene_taxonomy} are symbolic forms of
such constraints: they identify the quantities to compare and the physical
relation expected between them.

In the current implementation, most constraints are evaluated qualitatively or
ordinally rather than as exact calibrated equalities. For example, the verifier
checks whether successive rebound heights decrease during passive bouncing; it
does not estimate gravitational acceleration and require it to equal
\(9.81\,\mathrm{m/s^2}\). Monocular geometry, mask propagation, and acoustic-event
localization are not yet accurate enough for universal absolute thresholds.
Accordingly, the \texttt{decision\_basis} guides checks of temporal alignment,
ordering, monotonic trends, and qualitative consistency, and the verifier
returns support, violation, or uncertainty from the available quantities.

\subsection{Representative Physical Constraints}
\label{app:law-examples}

The Physical Law Selector operates under the constraints defined in
Section~\ref{sec:method}: each selected law must be relevant to the clip caption,
supported by the available physical quantities, and specified through a
structured schema containing its physical basis, operational check,
applicability, and failure conditions. Across completed runs, the selected laws
are operationalized through the recurring checks summarized below. The listed
names identify these checks in the \texttt{law\_name} fields of the output schema.
\begin{itemize}
  \item \textbf{Onset-to-event anchoring.} Examples include
  \path{impact_synchronization} for ball-table impacts,
  \path{release_onset_synchronization} for slingshot release, and
  \path{deflation_onset_visual_sync} for balloon collapse. These checks use
  \texttt{events[].onset\_ms} as the temporal anchor and search nearby frames for
  contact, release, rebound, relation-distance minima, velocity changes, or
  area-change events.
  \item \textbf{Kinematic and energy continuity.} Examples include
  \path{bounce_energy_dissipation_consistency},
  \path{passive_cart_energy_continuity}, and
  \path{post_release_trajectory_continuity}. These checks ask whether
  visual dynamics follow plausible gravity, inertia, dissipation, or projectile
  continuity, and whether the acoustic strength is compatible with the visible
  mechanical energy.
  \item \textbf{Material and acoustic envelope consistency.} Examples include
  \path{material_timbre_consistency} for ping-pong-ball impacts,
  \path{airflow_rubber_timbre_consistency} for balloon deflation, and
  \path{rolling_friction_envelope_coverage} for slope-descent scenes. These
  checks compare duration, decay, spectral novelty, and band energy with the
  material interaction implied by the video.
  \item \textbf{Source-gating and negative plausibility.} Examples include
  \path{per_event_unattributed_transient_coverage},
  \path{pre_contact_independent_onset_plausibility}, and
  \path{late_or_offscreen_unsourced_onset}. These checks flag strong
  or repeated audio onsets when the visual stream has no plausible source,
  contact, release, or remaining visible activity.
  \item \textbf{Scene-geometry and topology checks.} Examples include
  \path{static_reference_stability} for fixed ramps or supports,
  \path{topology_orientation_continuity} for balloon shape evolution, and
  \path{post_release_free_flight_continuity} for catapult projectiles.
  These checks identify geometry-level violations that local audio-visual
  synchronization alone may miss.
\end{itemize}

\subsection{Compute Cost}
\label{app:compute-cost}

Across 1,464 evaluated samples, complete \method{} inference uses 10,248
successful API calls. Each sample uses two captioning calls, one sampler-selector
call, one physical-law-selection call, and three physical-consistency-verification calls.
Normalized per sample, provider usage logs report 144.8k input tokens, 16.5k
output tokens, and 161.8k total tokens, with 18.1\% of the input served from
cache; token counts are rounded to 0.1k. Failed retries and extra successful
reruns are excluded. We report token consumption rather than dollar cost because
provider prices can change.

We estimate the input-token reduction from physical-quantity compression on a
stratified set of 135 clips, containing five clips for each scene--source
combination. For every clip, we replace only the compressed audio and video CSV
blocks with their original JSON quantity records. Provider usage logs show that the
compressed prompts average 8.9k input tokens, while the raw quantity prompts
average 154.3k input tokens, corresponding to about a 94\% reduction. As a
tokenizer-independent size check, the audio and video quantity blocks
decrease from 294.7k to 8.4k characters on average, a 97.1\% reduction. The
visual sampler retains 31.2 of 90.3 frames on average (34.5\%).

%% file: tables/ablation_results.tex
\begin{table}[!t]
\centering
\scriptsize
\caption{Physical-estimator ablations on the test split. Accuracy and F1 are reported as percentages.}
\label{tab:ablations}
\setlength{\tabcolsep}{3pt}
\begin{tabularx}{\columnwidth}{@{}lYY|YY@{}}
\toprule
Variant & \multicolumn{2}{c}{Seedance} & \multicolumn{2}{c}{VEO} \tabularnewline
\cmidrule(lr){2-3}\cmidrule(lr){4-5}
 & Acc(\%) & F1(\%) & Acc(\%) & F1(\%) \tabularnewline
\midrule
Full estimator & \textbf{70.30} & \textbf{64.29} & \textbf{72.16} & \textbf{65.82} \tabularnewline
w/o video enhancement & 60.40 & 54.55 & 62.89 & 57.14 \tabularnewline
w/o high-band branch & \underline{64.85} & \underline{60.34} & \underline{65.46} & \underline{59.88} \tabularnewline
\bottomrule
\end{tabularx}
\end{table}